\documentclass[letterpaper]{article} 
\usepackage{aaai2026}  
\usepackage{times}  
\usepackage{helvet}  
\usepackage{courier}  
\usepackage[hyphens]{url}  
\usepackage{graphicx} 
\usepackage{natbib}  
\usepackage{caption} 
\usepackage{algorithm}
\usepackage{algorithmic}
\usepackage{newtxmath}
\usepackage{amsmath}
\usepackage{booktabs}
\usepackage{multirow}
\usepackage[ruled,algo2e]{algorithm2e}
\usepackage{newfloat}
\usepackage{listings}
\usepackage{subfig}

\usepackage{xcolor}
\usepackage{times}
\usepackage{helvet}
\usepackage{courier}
\DeclareCaptionStyle{ruled}{labelfont=normalfont,labelsep=colon,strut=off} 
\floatstyle{ruled}
\newfloat{listing}{tb}{lst}{}
\floatname{listing}{Listing}
\title{Well Begun, Half Done: Reinforcement Learning with Prefix Optimization \\ for LLM Reasoning}
\author{
    Yiliu Sun\textsuperscript{\rm 1}, Zicheng Zhao\textsuperscript{\rm 1}, Yang Wei\textsuperscript{\rm 2}, Yanfang Zhang\textsuperscript{\rm 1}, Chen Gong\textsuperscript{\rm 3}\thanks{Corresponding author.}
}
\affiliations{
    \textsuperscript{\rm 1}School of Computer Science and Engineering, Nanjing University of Science and Technology, Nanjing, China.\\
    \textsuperscript{\rm 2}School of Software, North University of China, Taiyuan, China.\\
    \textsuperscript{\rm 3}School of Automation and Intelligent Sensing, Shanghai Jiao Tong University, Shanghai, China. \\
    chen.gong@sjtu.edu.cn
}

\usepackage{bibentry}

\begin{document}

\maketitle

\begin{abstract}
Reinforcement Learning with Verifiable Rewards (RLVR) significantly enhances the reasoning capability of Large Language Models (LLMs). Current RLVR approaches typically conduct training across all generated tokens, but neglect to explore which tokens (\textit{e.g.}, prefix tokens) actually contribute to reasoning. This uniform training strategy spends substantial effort on optimizing low-return tokens, which in turn impedes the potential improvement from high-return tokens and reduces overall training effectiveness. To address this issue, we propose a novel RLVR approach called Progressive Prefix-token Policy Optimization (PPPO), which highlights the significance of the prefix segment of generated outputs. Specifically, inspired by the well-established human thinking theory of Path Dependence, where early-stage thoughts substantially constrain subsequent thinking trajectory, we identify an analogous phenomenon in LLM reasoning termed Beginning Lock-in Effect (BLE). PPPO leverages this finding by focusing its optimization objective on the prefix reasoning process of LLMs. This targeted optimization strategy can positively influence subsequent reasoning processes, and ultimately improve final results. To improve the learning effectiveness of LLMs on how to start reasoning with high quality, PPPO introduces two training strategies: (a) Progressive Prefix Retention, which shapes a progressive learning process by increasing the proportion of retained prefix tokens during training; (b) Continuation Accumulated Reward, which mitigates reward bias by sampling multiple continuations for one prefix token sequence, and accumulating their scores as the reward signal. Extensive experimental results on various reasoning tasks (\textit{e.g.}, math, physics, chemistry, biology) demonstrate that our proposed PPPO outperforms representative RLVR methods, with the accuracy improvements of 18.02\% on only 26.17\% training tokens.
\end{abstract}


\section{Introduction}

Since the advance of Large Language Models (LLMs), reasoning has emerged as one of their core capabilities~\cite{lyu2025multi, wenjuan2025rulemaster+, talmor2019commonsenseqa, yureclor, qin2025dynamic, sun2025cortexdebate, zhao2023towards, zhang2025large, zhao2024graph, wang2025litesearch, sun2025fast, ding2025similar}, which demonstrates a promising path towards Artificial General Intelligence (AGI)~\cite{openai2025competitiveprogramminglargereasoning}. Therefore, it is critical to analyze and improve the reasoning capability of LLMs.

Recent advancements in LLMs, such as DeepSeek R1~\cite{guo2025deepseek} and OpenAI o1~\cite{openai2024openaio1card}, have demonstrated promising progress on various reasoning benchmarks. One of the key factors to drive these improvements is Reinforcement Learning with Verifiable Rewards (RLVR)~\cite{lambert2025tulu3pushingfrontiers, shao2024deepseekmath, xie2025logic, zheng2025act, zhang2025adaptthink, lou2025sequential}. Generally, RLVR adopts rule-based verification functions (\textit{e.g.}, correctness check~\cite{guo2025deepseek, yu2025dapo}, semantic entropy~\cite{zhao2025learningreasonexternalrewards}) to provide reward signals for sampled outputs, and then optimizes model outputs towards high quality by encouraging generations of the content with high reward signals.

However, current RLVR approaches face the challenge of limited training effectiveness. Specifically, most existing methods directly train across all generated tokens indiscriminately, but neglect the variable contributions of different tokens to the final performance. Given that tokens play heterogeneous roles in reasoning processes, this uniform treatment spends considerable effort on optimizing low-impact tokens, which in turn reduces training effectiveness and hinders further performance improvement brought by critical tokens~\cite{wang2025beyond, gu2024enhancing}.

\begin{figure*}[t]
\centering
\includegraphics[width=.95\textwidth]{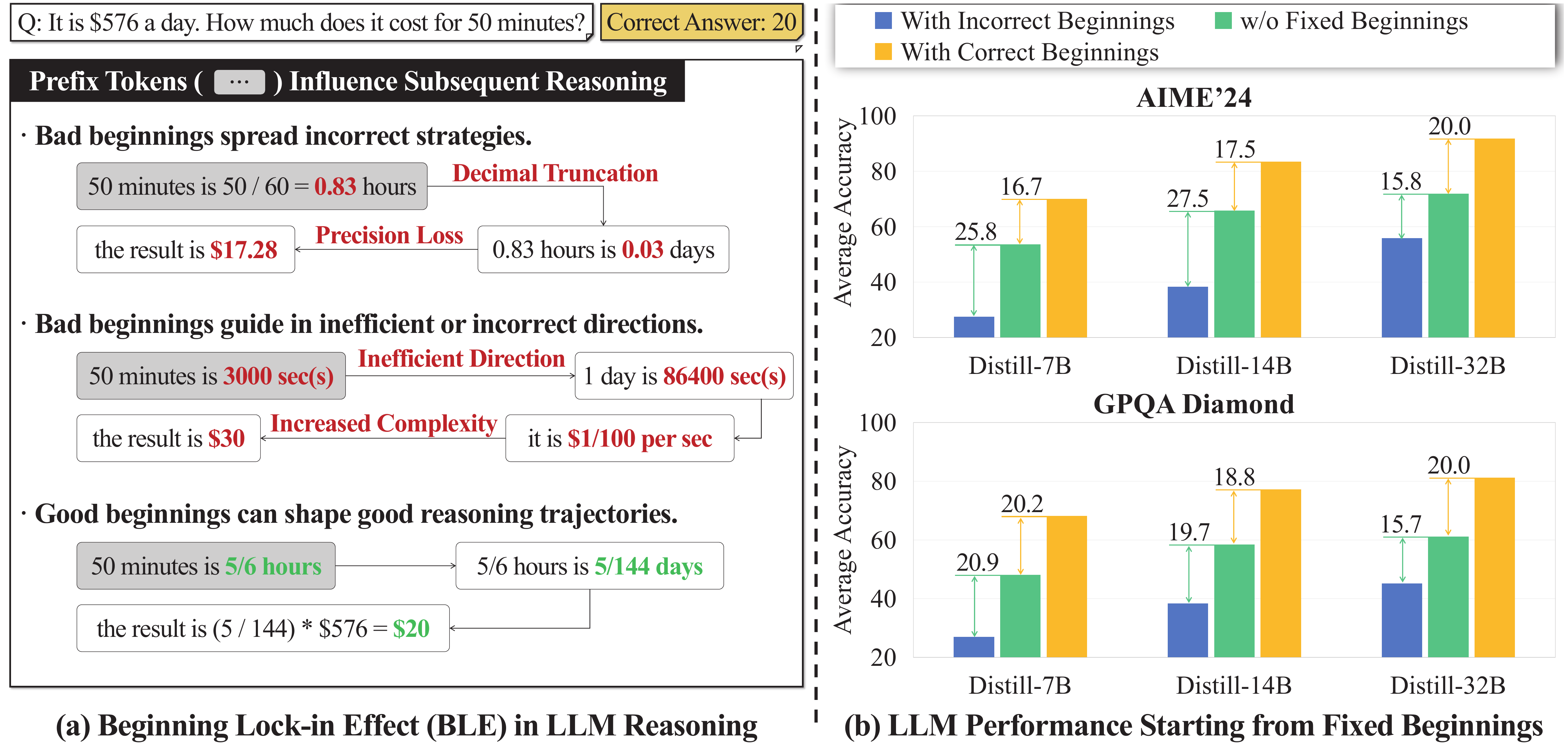} 
\caption{(a) In LLM reasoning, we identify a critical phenomenon termed BLE, where the initial reasoning significantly affects the subsequent reasoning. (b) Reasoning from prefix tokens extracted from correct outputs significantly improves the reasoning performance, while prefix tokens extracted from incorrect outputs lead to substantial performance degradation.}
\label{introduction_figure}
\end{figure*}

To address this issue, this paper proposes \textbf{P}rogressive \textbf{P}refix-token \textbf{P}olicy \textbf{O}ptimization (PPPO), a novel RLVR method inspired by the well-established human thinking theory of \textbf{Path Dependence}~\cite{david1975technical}. As revealed in~\cite{david1975technical}, during the human thinking process, the initial thought pattern substantially constrains the subsequent thinking trajectory, which in turn significantly influences the final result. This theory highlights the foundational influence of initial thoughts on the entire thinking process.

Given the autoregressive generation nature of LLMs, where each token is generated based on previous ones, we explore whether path dependence manifests in LLM reasoning (see the Preliminary section for details). Through extensive empirical investigations, we identify an analogous phenomenon termed \textbf{B}eginning \textbf{L}ock-in \textbf{E}ffect (BLE), where the initial reasoning process significantly constrains the subsequent reasoning steps and influences final results. For instance, as illustrated in Figure~\ref{introduction_figure}a, the flawed strategy adopted by the initial reasoning (\textit{e.g.}, decimal truncation or unnecessary unit conversion) misleads the reasoning process into an incorrect (precision loss) or inefficient (increased complexity) trajectory, and results in an incorrect result.

Based on this insight, our proposed PPPO improves the effectiveness of RLVR training by applying policy updates exclusively to prefix tokens (\textit{i.e.}, the first small part of an output token sequence) while masking the gradients of subsequent tokens. During the PPPO training, two key training strategies are incorporated, including \textbf{Progressive Prefix Retention} and \textbf{Continuation Accumulated Reward}. The former one forms a progressive learning process from simple to complex by gradually increasing the fraction of retained prefix tokens during PPPO training. The latter one requires the trained LLM to generate multiple continuations from a fixed prefix token sequence extracted from a full output, and then accumulates their scores as the reward signal for that prefix token sequence. By explicitly teaching LLMs to start reasoning with high quality, PPPO steers the entire reasoning trajectory toward better reasoning performance.

The effectiveness of PPPO has been well confirmed by extensive experiments on five complex reasoning benchmarks. For instance, when compared with the representative RLVR methods, PPPO increases the task accuracy by up to 18.02\% on a single benchmark (AIME'25), and increases the average accuracy across all the benchmarks by up to 14.64\%. Additionally, PPPO results in efficient reasoning, which reduces output tokens by up to 18.35\% while maintaining high accuracy. These experimental results demonstrate that our PPPO enables LLMs to start reasoning with high quality, which can lead to correct and concise reasoning trajectories.

The main contributions are summarized as follows:

\begin{itemize}
    \item We identify a critical phenomenon in LLM reasoning termed \textbf{Beginning Lock-in Effect (BLE)}, which demonstrates that the prefix tokens significantly shape subsequent reasoning trajectories.
    \item We propose a novel RLVR method termed \textbf{Progressive Prefix-token Policy Optimization (PPPO)}. Building upon BLE, PPPO can improve the training effectiveness of RLVR and enhance the reasoning capability of LLMs by focusing on the optimization of prefix tokens.
    \item We introduce two key training strategies to improve the learning effectiveness of LLMs during PPPO training, including \textbf{Progressive Prefix Retention} and \textbf{Continuation Accumulated Reward}.
    \item We conduct extensive experiments on various reasoning benchmarks to show that \textbf{our PPPO consistently outperforms representative RLVR baseline methods} (\textit{i.e.}, GRPO, DAPO, INTUITOR, and DAPO-FT).
\end{itemize}

\section{Related Work}

Reinforcement learning with verifiable rewards~\cite{qu2025latent, rafailov2023direct, byun2024ares, he2024planning, zhang2024rest} has been a promising framework to enhance the reasoning capability of LLMs. Typically, RLVR fine-tunes LLMs by using reward signals derived from human preferences or automatic task-specific validations to improve the quality of LLM outputs. For example, \citet{schulman2017proximal} introduce Proximal Policy Optimization (PPO), which adopts Kullback-Leibler Divergence to constrain the magnitude of parameter updates and improve the training stability. Notably, PPO needs to train a critic model to estimate the advantages of LLM outputs, which is costly. Hence, \citet{shao2024deepseekmath} propose Group Relative Policy Optimization (GRPO), which foregoes the critic model and estimates the advantage of each output according to the average score of all the outputs instead. Encouraged by the success of GRPO, many RLVR methods that do not require these costly models have been proposed~\cite{yeo2025demystifying, xie2025logic, zheng2025act, xu2025redstar, yue2025vapo, zhang2025adaptthink}. For instance, \citet{yu2025dapo} introduce Dynamic Sampling Policy Optimization (DAPO) to alleviate four issues (\textit{i.e.}, entropy collapse, training instability, low training efficiency, and reward hacking) faced by GRPO. \citet{zheng2025act} propose GRPO with Efficient Selective Rollout (GRESO), which improves the quality of sampled problems and thus enhances the training efficiency.

However, existing RLVR methods suffer from low training effectiveness~\cite{sheng2025hybridflow, wang2025beyond, zheng2025act}. Specifically, these methods train across all tokens, but overlook the heterogeneous contributions of different tokens to reasoning performance. As a result, much effort is spent on optimizing tokens with low or negative returns. Although \citet{wang2025beyond} attempt to address this issue by solely optimizing high-entropy tokens (\textit{e.g.}, wait, however, rethink), this strategy encourages LLMs to explore, but cannot guarantee the quality of the reasoning paths triggered by these high-entropy tokens. As a result, the improvement brought by this strategy is unstable.

Distinctively, inspired by the human thinking theory of Path Dependence, our PPPO improves the training effectiveness of RLVR by focusing on optimizing prefix tokens, which play critical roles in shaping the reasoning trajectory. Hence, by learning to start reasoning effectively, LLMs can maintain a coherent and high-quality reasoning trajectory, which in turn improves the final performance.

\section{Preliminaries}

In this section, we first introduce the concept of Path Dependence in human thinking processes and then explore its counterpart in the reasoning processes of LLMs.

\subsection{Path Dependence}

Path dependence~\cite{david1975technical} is a well-established concept of human thinking patterns, where early-stage thoughts constrain subsequent thoughts. Based on the effects on subsequent thoughts, it can be categorized as either benign or malignant. A benign one means that high-quality initial thoughts steer the thinking process toward a desirable outcome, while a malignant one indicates that low-quality initial thoughts result in a suboptimal outcome. Namely, initial thoughts significantly influence long-term outcomes, which suggests that high-quality initial thoughts increase the likelihood of success~\cite{page2006path}.

\begin{figure}[t]
\centering
\subfloat[LLMs struggle to escape from the trajectories shaped by incorrect prefix token sequences.]{
		\includegraphics[width=.93\columnwidth]{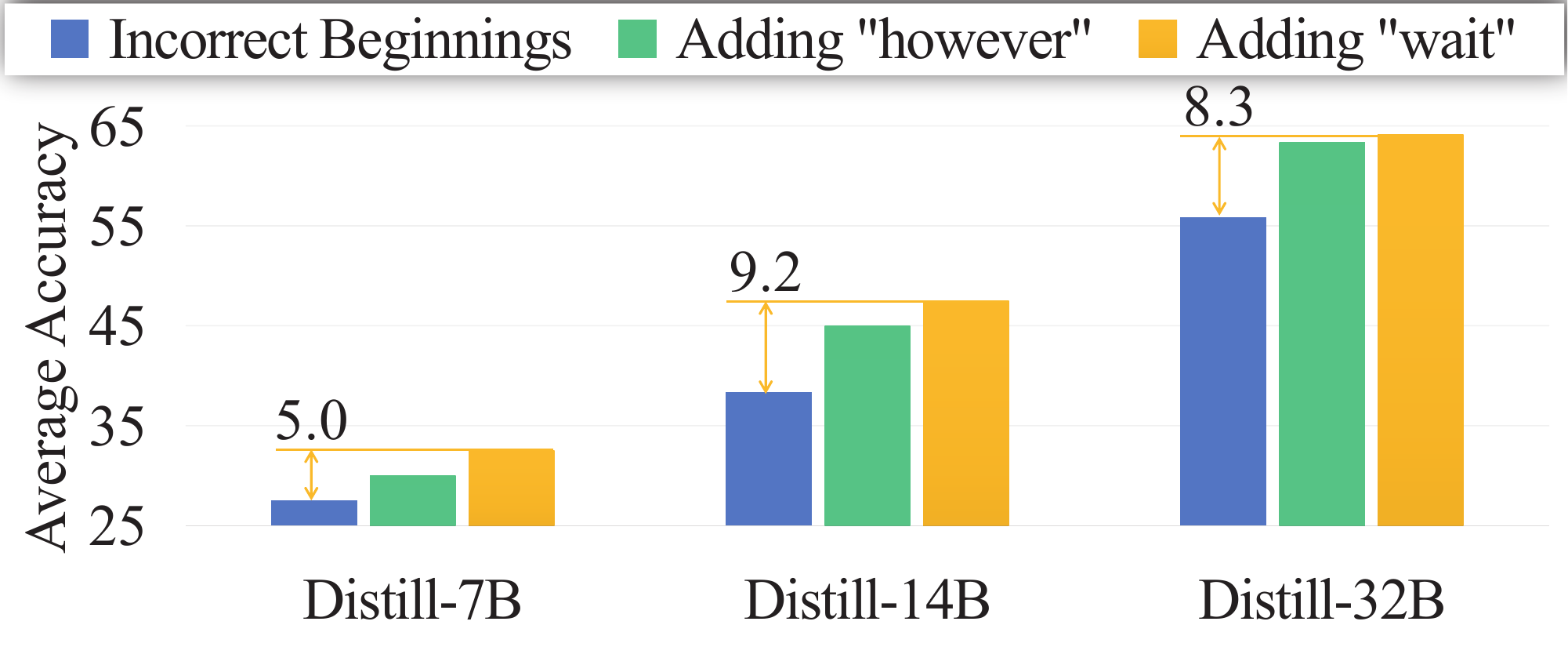}
        \label{fig_a}}\\
\subfloat[As the length of the prefix token sequence increases, BLE gradually establishes.]{
		\includegraphics[width=.93\columnwidth]{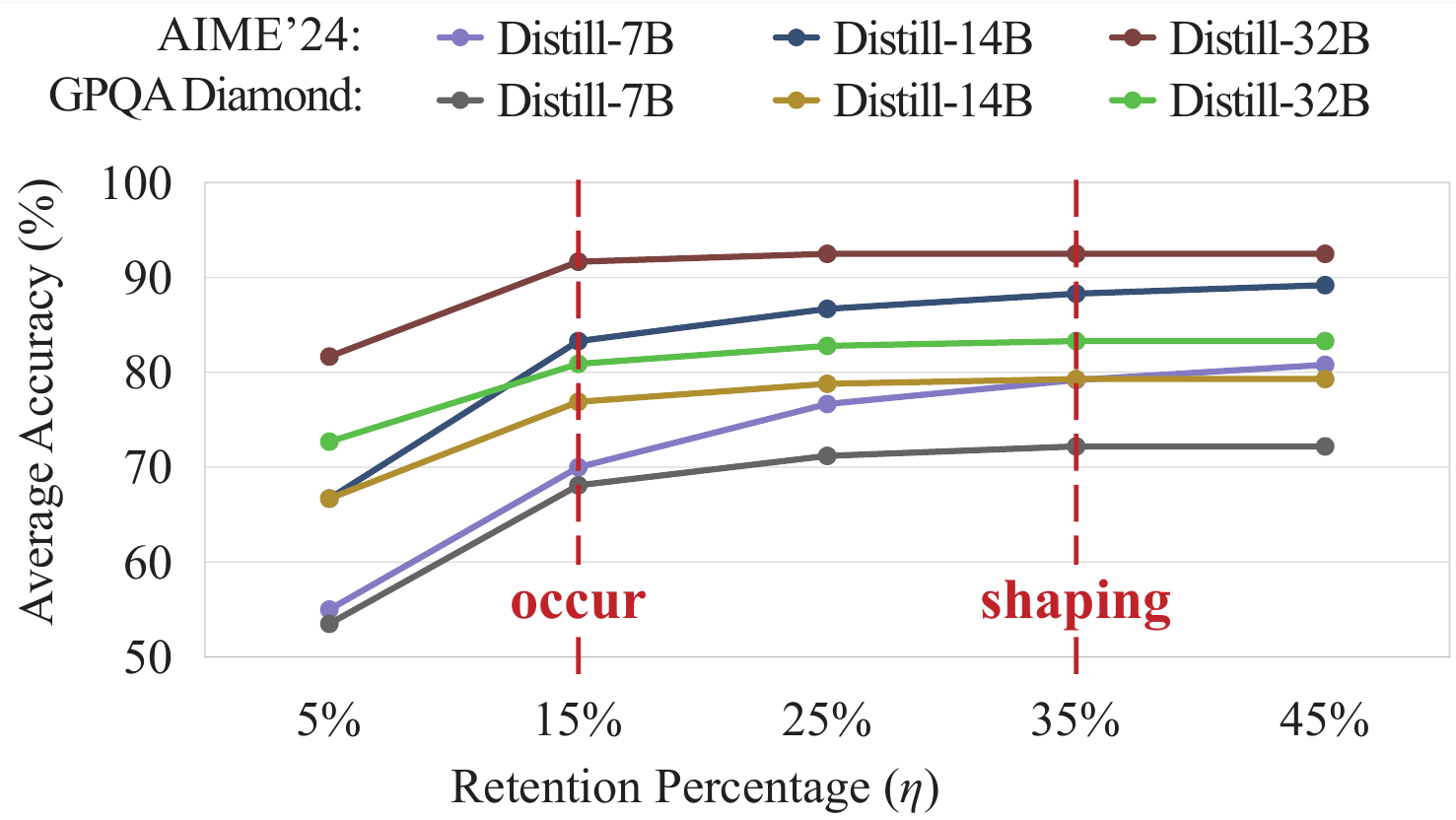}
        \label{fig_b}}
\caption{Explorations of beginning lock-in effect.}
\label{preli_main3}
\end{figure}

\begin{figure*}[t]
\centering
\includegraphics[width=.95\textwidth]{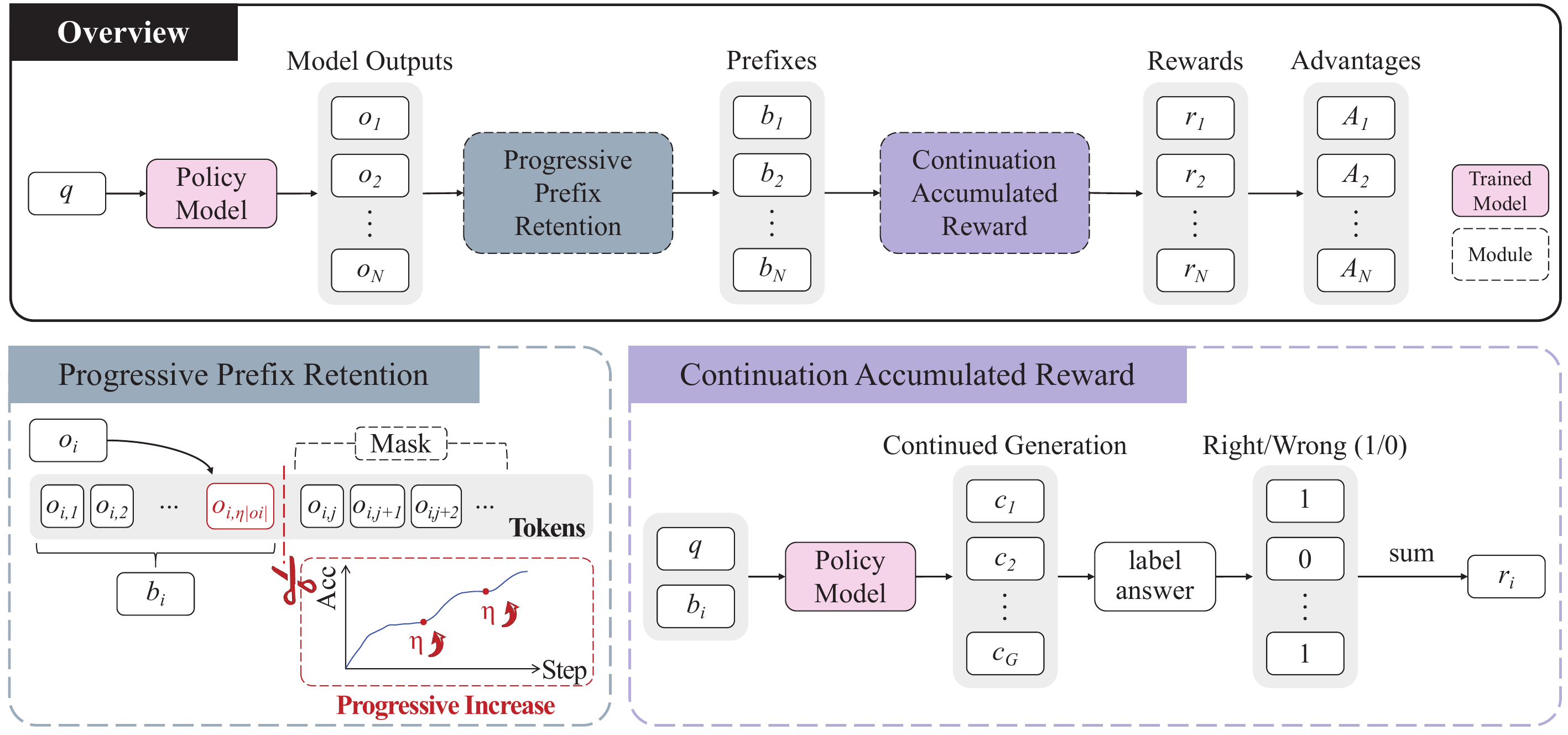} 
\caption{Illustration of PPPO, which is inspired by the human thinking theory of Path Dependence. PPPO focuses on optimizing the prefix segments of LLM outputs and thus positively influences the subsequent reasoning paths. During the PPPO training, two strategies are introduced, including ``Progressive Prefix Retention'' and ``Continuation Accumulated Reward''.}
\label{PPPO}
\end{figure*}

\subsection{Beginning Lock-in Effect}

Due to the autoregressive nature of LLMs, where each token is generated based on previous tokens, a critical research question is raised: \textit{Does the quality of prefix tokens exert a significant influence on the final result in LLM reasoning?}

To answer this question, we design controlled experiments, in which LLMs are required to reason from fixed prefix tokens. Specifically, for each problem in two reasoning benchmarks ( \textit{i.e.}, AIME'24~\cite{codeforcesamerican} and GPQA Diamond~\cite{rein2024gpqa}), we employ DeepSeek-R1-Distill-Qwen series~\cite{guo2025deepseek} to generate 4 correct and incorrect outputs, respectively. We then extract the first 15\% of the tokens from each output as a fixed prefix token sequence. For each fixed prefix token sequence, we prompt the original LLM to generate eight continuations and calculate the average accuracy. As reported in Figure~\ref{introduction_figure}b, we observe that all the evaluated LLMs exhibit a significant performance decline (up to 27.5\% on AIME'24 with DeepSeek-R1-Distill-Qwen-14B) when reasoning from the fixed prefix token sequences sampled from incorrect answers, while demonstrating substantial improvement (up to 20.2\%) when initialized with the fixed prefix token sequences from correct answers. 

To test whether LLMs can recover from the low-quality initial reasoning, we add ``wait'' and ``however'' after those malignant prefix tokens, which can encourage LLMs to reflect previous reasoning~\cite{muennighoff2025s1simpletesttimescaling, wang20258020rulehighentropyminority}. As shown in Figure~\ref{fig_a}, even with these corrective interventions, the maximum recovery of accuracy reaches only 9.2\%. The results reveal that LLMs struggle to escape from the flawed initial reasoning, and further support the path dependence hypothesis in LLM reasoning. We term this phenomenon as \textbf{B}eginning \textbf{L}ock-in \textbf{E}ffect (BLE).

\section{Methodology}

This section introduces the overall framework of our \textbf{P}rogressive \textbf{P}refix-token \textbf{P}olicy \textbf{O}ptimization (PPPO). As described in the Preliminaries section, high-quality initial reasoning paths yield high-quality results, while low-quality initial reasoning paths harm model performance. This phenomenon (BLE) can be expressed as follows:

\begin{flalign}
P &\left ( \hat{y} = a \mid q , I_{high-quality}  \right ) \nonumber \\
& > P\left ( \hat{y} = a \mid q \right ) \nonumber \\
& > P\left ( \hat{y} = a \mid q , I_{low-quality}  \right ),
\end{flalign}

\noindent where $\hat{y}$, $q$, $a$, and $I$ denote an answer of the LLM, a question, the correct answer, and an initial reasoning path, respectively. Inspired by BLE, PPPO enhances LLM reasoning by explicitly optimizing the initial stages of reasoning processes. As illustrated in Figure~\ref{PPPO}, PPPO retains the gradients of the first small portion of generated tokens, and masks those of subsequent tokens. This strategy enables LLMs to focus on learning high-quality reasoning initiations, which in turn improves the subsequent reasoning performance.

Specifically, for $q$ and its correct answer $a$ in the dataset $\mathcal{D}$, PPPO samples $N$ outputs $\left \{ \mathbf{o}_{i}  \right \} _{i= 1}^{N} $ from the policy model $\pi _{\theta}$, and retains their first $\eta$ of tokens to obtain $N$ prefix sequences $\left \{ \mathbf{b}_{i}  \right \} _{i= 1}^{N} $, where $\eta$ is a percentage. Starting from each $\mathbf{b}_{i}$, PPPO samples $G$ outputs $\{\mathbf{c}_{j}\}_{j=1}^{G}$. Subsequently, PPPO optimizes the policy via the following objective:

\begin{flalign}
\label{EQPPPO}
 \mathcal{J} _{PPPO} & \left ( \theta  \right ) = \mathbb{E} _{\left ( q,a \right )\sim \mathcal{D}, \left \{ \mathbf{o}_{i}  \right \}_{i=1}^{N}\sim \pi _{\theta _{old}\left ( \cdot \mid q \right ) }, \left \{ \mathbf{c}_{j}  \right \}_{j=1}^{G}\sim \pi _{\theta _{old}\left ( \cdot \mid q,\mathbf{b}_{i} \right ) } }  \nonumber \\
& \Bigg [ \frac{1}{ {\textstyle \sum_{k=1}^{N} \left | \mathbf{o}_{k}  \right | } } \sum_{i=1}^{N} \sum_{j=1}^{ \left | \mathbf{o}_{i}  \right | } \mathrm{H} \left ( j,\mathbf{o}_{i}  \right ) \cdot  \min \left ( r_{i,j} \left ( \theta  \right ) \hat{A}_{i,j} ,  \right.   \nonumber \\
 &  \left. \mathrm{clip} \left ( r_{i,j} \left ( \theta  \right ) , 1- \varepsilon _{low}, 1+  \varepsilon _{high}   \right )   \hat{A}_{i,j}   \right )  \Bigg ],
\end{flalign}

\noindent where

\begin{flalign}
r_{i,j}\left ( \theta  \right )  = \frac{\pi _{\theta } \left ( o_{i,j} \mid q,\mathbf{o}_{i,< j}  \right ) }{\pi _{\theta _{old} } \left ( o_{i,j} \mid q, \mathbf{o}_{i,< j}  \right ) } ,
\end{flalign}

\begin{flalign}
\hat{A}_{i,j} =\frac{R_{i} -\mathrm{mean} \left ( \left \{ R_{i}   \right \} _{i=1}^{N}  \right ) }{\mathrm{std} \left ( \left \{ R_{i}   \right \} _{i=1}^{N}  \right ) } .
\end{flalign}

\noindent In above equations, $\mathrm{H} \left( \cdot, \cdot \right)$ retrains the advantages of prefix tokens which will be detailed in Eq.~\eqref{H}, $\mathrm{clip} \left( \cdot, \cdot, \cdot \right )$ denotes a constraint function for policy update, $\mathrm{mean} \left( \cdot \right )$ denotes a function which calculates the average value, $\mathrm{std} \left( \cdot \right )$ denotes a function which calculates the standard deviation, $\left | \mathbf{o}_{i} \right |$ returns the token number of $\mathbf{o}_i$, $\varepsilon$ is the clipping range of the importance sampling ratio, and $R_i$ is the reward of $\mathbf{o}_i$. The algorithm can be found in Algorithm~\ref{alg1}. To enhance the learning effectiveness on starting reasoning with high quality, PPPO incorporates two key strategies, including Progressive Prefix Retention and Continuation Accumulated Reward, which are described in the following subsections.

\begin{algorithm2e}[t]
\caption{Training Process of PPPO}
\label{alg1}
\KwIn{Initial policy model $\pi_{\theta}$; reward model $R$; task set $\mathcal{D}$; hyperparameters $\varepsilon_{low}$, $\varepsilon_{high}$; initial sequence proportion $\eta$; maximum training steps $S$}
\KwOut{Optimized policy model $\pi_{\theta}$}
\For{step $= 1,2,\dots, S$}{
    Sample a batch of tasks $\mathcal{D}_d \sim \mathcal{D}$\;
    Set old policy parameters $\pi_{\theta_{old}} \gets \pi_{\theta}$\;
    \For{\rm{each query} $q \in \mathcal{D}_d$}{
        Sample $N$ outputs $\{\mathbf{o}_i\}_{i=1}^{N}$ from policy $\pi_{\theta}$\;
        Retain the first $\eta$ tokens of each output to form prefix token sequences $\{\mathbf{b}_{i}\}_{i=1}^{N}$\;
        \For{\rm{each sequence} $\mathbf{b}_{i} \in \{\mathbf{b}_{i}\}_{i=1}^{N}$}{
            Sample $\{\mathbf{c}_j\}_{j=1}^{G}$ starting from $q$ and $\mathbf{b}_{i}$\;
        }
        Compute rewards $\{\mathbf{r}_i\}_{i=1}^{N}$ for $\{\mathbf{b}_{i}\}_{i=1}^{N}$ using reward model $R$\;
    }
    \For{\rm{each sequence} $\mathbf{b}_{i}$}{
        Compute the advantage $\hat{A}_{i,j}$ for each token $b_{i,j}$ within sequence $\mathbf{b}_{i}$\;
    }
    Update $\theta$ by using calculated advantages $\mathbf{\hat{A}}$\;
    \If{\rm{performance improvement of} $\pi_{\theta}$ \rm{reaches a bottleneck}}{
        Increase $\eta$ for subsequent iterations\;
    }
}
\Return Optimized policy model $\pi_{\theta}$\;
\end{algorithm2e}

\subsection{Progressive Prefix Retention}

For each $\mathbf{o}_{i}$, PPPO retains the gradients of the first $\eta$ tokens for optimization and masks those of the remaining tokens, which can be achieved by $\mathrm{H} \left( \cdot, \cdot \right)$:

\begin{flalign}
\label{H}
\mathrm{H} \left ( j,\mathbf{o}_{i}  \right ) = \vmathbb{1} \left ( j\le \left \lfloor \eta \cdot \left | \mathbf{o}_{i}  \right |  \right \rfloor \right ) ,
\end{flalign}

\noindent where $\vmathbb{1} \left( \cdot \right)$ and $\left \lfloor \cdot \right \rfloor$ denote an indicator function and a function which rounds a value down to the nearest integer, respectively. During the training, PPPO gradually increases $\eta$. The increasing trigger occurs when the LLM performance stagnates, which is defined as the absence of the accuracy increase on the validation set after multiple training steps:

\begin{flalign}
\eta = \begin{cases}
  \ \ \eta, & \text{ if } \Delta acc > 0  \\
  \ \ \eta + \Delta \eta, & \text{ if } \Delta acc \le  0
\end{cases},
\end{flalign}

\noindent where $\Delta \text{acc}$ and $\Delta \eta$ denote the accuracy change and the increment step of $\eta$, respectively. This strategy forms a progressive learning process from simple to complex. Specifically, PPPO starts with short prefix token sequences, which reduces the learning complexity and allows LLMs to rapidly develop core competencies of starting reasoning with high quality. Subsequently, PPPO gradually increases the optimized sequence length. Since the model has built a solid foundation before, it can maintain high learning quality and stability after increasing the optimized sequence length.

\subsection{Continuation Accumulated Reward}

Most existing RLVR methods use single-sample reward signals for output optimization, such as correctness verifications and format checks:

\begin{flalign}
R_{i}^{correct} = \vmathbb{1} \left ( \hat{y}_{\mathbf{o}_{i} }   = a \right ), \ \
R_{i}^{format} = \vmathbb{1} \left ( str \subseteq \mathbf{o}_{i} \right ),
\end{flalign}

\noindent where $\hat{y}_{\mathbf{o}_{i} }$ denotes the answer extracted from $\mathbf{o}_{i}$ and $str$ donates a string representing a specific reasoning structure. However, for prefix token sequences, the single-sample evaluation suffers from high stochasticity, which leads to biased reward signals and unstable policy updates. 

To alleviate this problem, PPPO introduces Continuation Accumulated Reward to evaluate prefix tokens. Specifically, for each $\mathbf{o}_{i} = (o_{i,1}, o_{i,2}, \dots, o_{i,\left | \mathbf{o}_{i}  \right | })$, PPPO retains the first $\eta$ of tokens to form $\mathbf{b}_{i}$:

\begin{flalign}
\mathbf{b}_{i} =\left \{ o_{i,j}  \right \}_{j=1}^{\left \lfloor \eta\cdot \left | \mathbf{o}_{i}  \right |  \right \rfloor }.
\end{flalign}

\noindent For each $\mathbf{b}_{i}$, PPPO samples $G$ continuations $\{\mathbf{c}_{j}\}_{j=1}^{G}$ conditioned on $\mathbf{b}_{i}$ and the corresponding question $q$:

\begin{flalign}
P\left ( \mathbf{c}_{j} \mid q,\mathbf{b}_{i}  \right ) = {\textstyle \prod_{k}^{}P\left ( c_{j,k} \mid \mathbf{c}_{j,< k} ,q,\mathbf{b}_{i}  \right ) } .
\end{flalign}

\noindent The reward $R_{i}$ for $\mathbf{b}_{i}$ is computed as the cumulative score of the continuations $\left \{ \mathbf{c}_{j}  \right \} _{j= 1}^{G}$ and the original output $\mathbf{o}_{i}$:

\begin{flalign}
R_{i}= {\textstyle \sum_{j=1}^{G} \vmathbb{1} \left ( \hat{y}_{\mathbf{c}_{j} }  = a \right )  } + \vmathbb{1} \left ( \hat{y}_{\mathbf{o}_{i} }   = a \right ) .
\end{flalign}

\noindent This accumulated reward mechanism mitigates randomness and provides reliable evaluations for prefix tokens.

\begin{table*}[t]
\centering
\begin{tabular}{lcccccc}
\toprule
\multicolumn{1}{c}{Method} & AIME'24 & AIME'25 & MATH 500 & AMC'23  & GPQA Diamond & Average \\ \midrule
Qwen3-1.7B    & 25.83 & 18.33 & 76.05  & 56.93 & 28.16      & 41.06 \\
\ \ \ + GRPO~\cite{guo2025deepseek}        & 27.50 & 20.00 & 82.10  & 60.24 & 29.80      & 43.93 \\
\ \ \ + DAPO~\cite{yu2025dapo}        & 30.83 & 23.33 & 83.60  & \textbf{63.03} & \underline{31.82}      & 46.52 \\
\ \ \ + INTUITOR~\cite{zhao2025learningreasonexternalrewards}    & 27.29 & 17.71 & 78.13  & 60.99 & 29.96      & 42.82 \\
\ \ \ + DAPO-FT~\cite{wang2025beyond}     & \underline{31.25} & \underline{23.96} & \underline{83.93}  & 62.80 & 31.44      & \underline{46.68} \\
\ \ \ + PPPO \textbf{(ours)} & \textbf{38.65}  & \textbf{28.96}  & \textbf{87.80}   & \underline{62.89}  & \textbf{39.52}       & \textbf{51.58}  \\ \midrule
Qwen3-4B      & 48.75 & 35.42 & 84.46  & 72.67 & 43.59      & 56.98 \\
\ \ \ + GRPO~\cite{guo2025deepseek}        & 52.08 & 37.71 & 88.40  & 76.77 & 46.78      & 60.35 \\
\ \ \ + DAPO~\cite{yu2025dapo}        & \underline{56.46} & \underline{42.08} & 92.33  & 81.63 & \underline{49.37}      & 64.37 \\
\ \ \ + INTUITOR~\cite{zhao2025learningreasonexternalrewards}    & 51.04 & 35.42 & 88.26  & 75.83 & 46.43      & 59.40 \\
\ \ \ + DAPO-FT~\cite{wang2025beyond}     & 56.25 & \underline{42.08} & \underline{92.38}  & \underline{82.00} & 49.21      & \underline{64.38} \\
\ \ \ + PPPO \textbf{(ours)} & \textbf{63.54}  & \textbf{53.44}  & \textbf{94.60}   & \textbf{83.06}  & \textbf{52.07}       & \textbf{69.34}  \\ \midrule
Qwen3-8B      & 52.29 & 38.75 & 86.06  & 75.08 & 46.12      & 59.66 \\
\ \ \ + GRPO~\cite{guo2025deepseek}        & 58.75 & 42.29 & 91.00  & 79.44 & 50.51      & 64.40 \\
\ \ \ + DAPO~\cite{yu2025dapo}        & 63.13 & 48.75 & 93.21  & 83.96 & \underline{55.18}      & 68.85 \\
\ \ \ + INTUITOR~\cite{zhao2025learningreasonexternalrewards}    & 55.42 & 40.83 & 91.20  & 78.24 & 49.31      & 63.00 \\
\ \ \ + DAPO-FT~\cite{wang2025beyond}     & \underline{63.75} & \underline{49.38} & \underline{93.65}  & \underline{84.11} & 54.77      & \underline{69.13} \\
\ \ \ + PPPO \textbf{(ours)} & \textbf{72.19}  & \textbf{59.69}  & \textbf{94.73}  & \textbf{86.75}  & \textbf{58.13}  & \textbf{74.30}  \\ \bottomrule
\end{tabular}
\caption{\label{main_table}
   Comparison results on 5 benchmarks. The unit of all the results is ``$\%$''. The best and second-best scores are shown in bold and underlined, respectively.
  }
\end{table*}

\section{Experiments}
In this section, we introduce the experimental setup and analyses of our experimental results.

\subsection{Experimental Setup}
In this part, we provide the experimental setup, including training details, baseline methods, and evaluation strategies.

\textbf{Training Details.} We adopt the Qwen3 series (~\textit{i.e.}, Qwen3-1.7B, Qwen3-4B and Qwen3-8B)~\cite{qwen3technicalreport} under the thinking mode as backbone models, and DAPO-Math-17K as the training dataset. For each question $q$, we sample 8 outputs $\left \{ \mathbf{o}_{i}  \right \} _{i=1}^{8} $. For each prefix token sequence $\mathbf{b}_{i}$, we sample 8 continuations $\{\mathbf{c}_{j}\}_{j=1}^{8}$. Additionally, the models are trained with a learning rate of $1 \times 10^{-6}$. By following~\cite{yu2025dapo}, we set $\varepsilon _{low}$ to 0.2 and $\varepsilon _{high}$ to 0.28.

Moreover, as shown in Figure~\ref{fig_b}, model performance exhibits a significant transition when the retention proportion $\eta$ reaches 15\%, and stabilizes when $\eta$ reaches 35\%. This case indicates that BLE manifests within the first 15\% of tokens, and essentially establishes by 35\%. Accordingly, the retention proportion $\eta$ in PPPO is initialized at 15\% and gradually increased to 35\% by steps of 5\%.

\textbf{Baseline Methods.} We compare our PPPO with 5 methods: (a) Backbone Models, (b) GRPO~\cite{shao2024deepseekmath}, (c) DAPO~\cite{yu2025dapo}, (d) INTUITOR~\cite{zhao2025learningreasonexternalrewards}, and (e) DAPO with Forking Tokens (DAPO-FT)~\cite{wang2025beyond}. For fair comparison, the maximum response length of LLMs is 10240 tokens in all the methods. Moreover, for each training step, PPPO samples $8 \times 8$ outputs and all the baseline methods sample 64 outputs.

\textbf{Evaluation Strategies.} We evaluate our method and the baseline methods on 5 benchmarks, which are widely used to evaluate the reasoning capability of LLMs: (a) AIME'24~\cite{codeforcesamerican}, (b) AIME'25~\cite{balunović2025matharenaevaluatingllmsuncontaminated}, (c) MATH 500~\cite{hendrycks2measuring}, (d) AMC'23~\cite{amc2023}, and (e) GPQA Diamond~\cite{rein2024gpqa}. By following~\cite{yue2025vapo, wang2025beyond}, for each dataset, we test 32 times under a zero-shot setting and report the average accuracy, \textit{avg@32}.

\subsection{Main Results}
In this part, we present the experimental results and detailed analyses to highlight the effectiveness of our PPPO.

\textbf{PPPO outperforms baseline methods.} As shown in Table~\ref{main_table}, PPPO achieves the highest accuracy in 14 out of 15 settings (3 backbone models $\times$ 5 benchmarks). The only exception is on the AMC'23 benchmark with the Qwen3-1.7B model, where PPPO slightly underperforms DAPO by 0.14\%. In all other cases, PPPO outperforms all the baseline methods (increasing accuracy by up to 18.02\%). These results demonstrate the consistent superiority of our proposed PPPO in enhancing the reasoning capability of LLMs. Additionally, we define the metric of LE as the ratio of Average Accuracy Increase (AAI) to Proportion of Optimized Tokens (POT) during RLVR training. As reported in Table~\ref{main_ex_3}, PPPO achieves the highest LE compared with all the baseline methods, which delivers accuracy improvements of up to 14.64\% while optimizing only 24.83\% of the generated tokens. The results reveal that PPPO enables LLMs to learn key reasoning abilities (\textit{e.g.}, high-quality reasoning initiation, effective reasoning strategies) from a few tokens, which further confirm the superior training effectiveness of PPPO.

\begin{figure}[t]
\centering
\includegraphics[width=.95\linewidth]{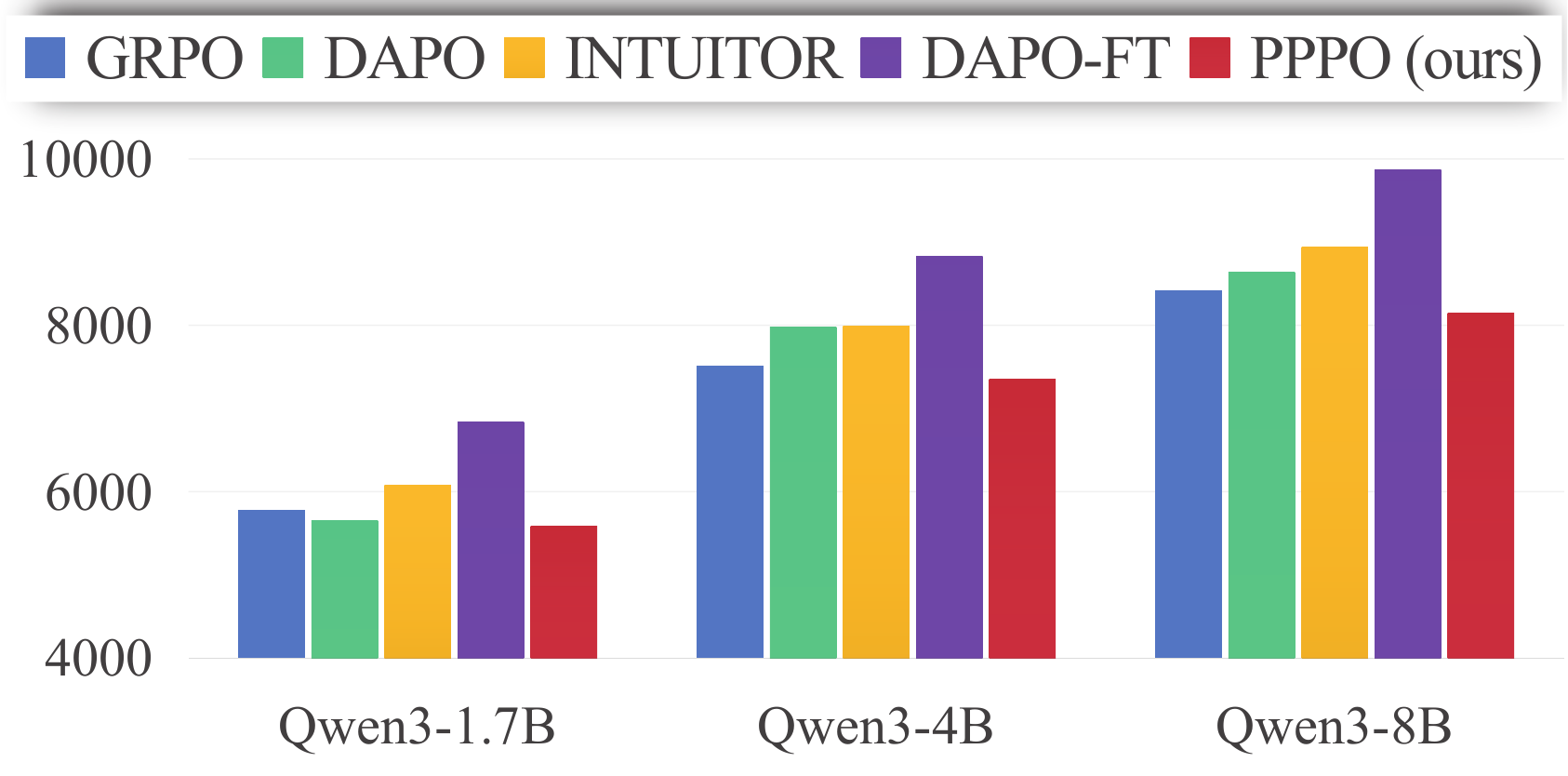} 
\caption{Average output lengths of the backbone models after being trained on each baseline method.}
\label{main_3}
\end{figure}

\textbf{PPPO results in efficient reasoning.} For each backbone model, we calculate the average token number of its outputs after being trained on each baseline method, and present the results in Figure~\ref{main_3}. While achieving the highest accuracies, PPPO achieves the lowest reasoning cost compared with other baseline methods, which reduces the average number of generated tokens by up to 18.35\%. The reason is that PPPO enables LLMs to start reasoning with high quality, which leads to concise yet effective reasoning trajectories.

\begin{table}[t]
\setlength{\tabcolsep}{2.5mm}
\centering
\begin{tabular}{c|cccc}
\toprule
Qwen3                 & Method      & AAI $\uparrow$    & POT $\downarrow$           & LE $\uparrow$
\\ \midrule
\multirow{5}{*}{1.7B} & GRPO        & 2.87              & 100.00         & 2.87          \\
                      & DAPO        & 5.46              & 100.00         & 5.46          \\
                      & INTUITOR    & 1.76              & 100.00         & 1.76          \\
                      & DAPO-FT     & \underline{5.62}              & \textbf{20.00}          & \underline{28.08}         \\
                      & PPPO (\textbf{ours}) & \textbf{10.52}             & \underline{29.19}          & \textbf{36.05}         \\ \midrule
\multirow{5}{*}{4B}   & GRPO        & 3.37              & 100.00         & 3.37          \\
                      & DAPO        & \underline{7.39}              & 100.00         & 7.39          \\
                      & INTUITOR    & 2.42              & 100.00         & 2.42          \\
                      & DAPO-FT     & \underline{7.39}              & \textbf{20.00}          & \underline{37.02}         \\
                      & PPPO (\textbf{ours}) & \textbf{12.36}             & \underline{26.17}          & \textbf{47.24}         \\ \midrule
\multirow{5}{*}{8B}   & GRPO        & 4.74              & 100.00         & 4.74          \\
                      & DAPO        & 9.19              & 100.00         & 9.19          \\
                      & INTUITOR    & 3.34              & 100.00         & 3.34          \\
                      & DAPO-FT     & \underline{9.47}              & \textbf{20.00}          & \underline{47.36}         \\
                      & PPPO (\textbf{ours}) & \textbf{14.64}             & \underline{24.83}          & \textbf{58.95}   \\ \bottomrule     
\end{tabular}
\caption{\label{main_ex_3}
   Comparison of training effectiveness. Here, ``$\uparrow$'' means that higher values are better, and ``$\downarrow$'' means that lower values are better. The best and second-best scores are shown in bold and underlined, respectively.
  }
\end{table}

\subsection{Performance Investigation}
In this part, we conduct in-depth investigations on our proposed PPPO to analyze its effectiveness.

\begin{figure}[t]
\centering
\includegraphics[width=.95\linewidth]{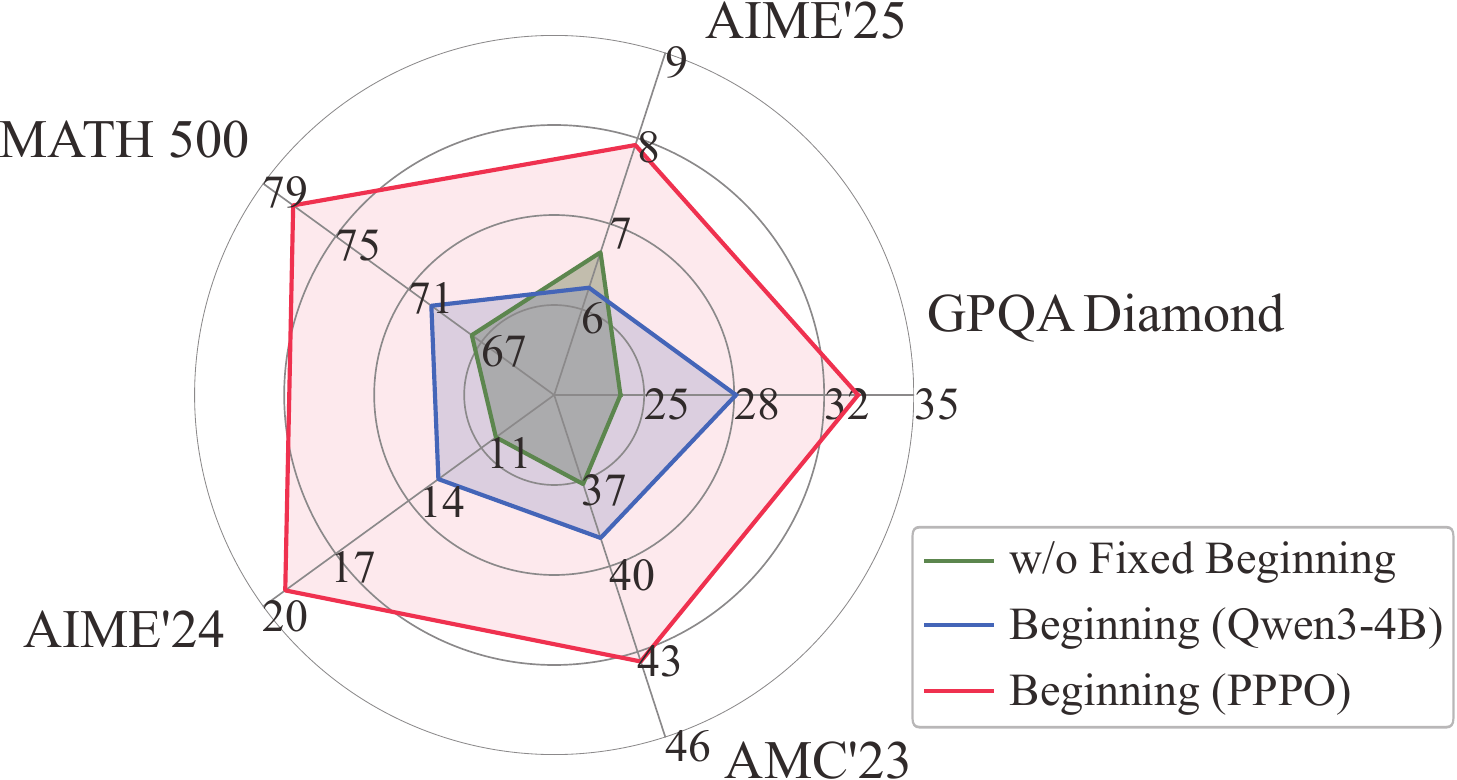} 
\caption{Task performance of Qwen2.5-7B-Instruct when reasoning from prefix tokens extracted from Qwen3-4B with and without PPPO training.}
\label{performance_1}
\end{figure}

\textbf{PPPO enables high-quality reasoning initiation.} We require Qwen2.5-7B-Instruct~\cite{qwen2.5} to reason starting from prefix tokens extracted from Qwen3-4B with and without PPPO training. As shown in Figure~\ref{performance_1}, the prefix tokens extracted from untrained Qwen3-4B provide only marginal performance gains for Qwen2.5-7B-Instruct, and even reduce accuracy on AIME'25. In contrast, starting reasoning from the prefix tokens extracted from Qwen3-4B with PPPO training enables Qwen2.5-7B-Instruct to achieve optimal performance across all benchmarks, which improves accuracy by up to 9.04\%. The results confirm that our proposed PPPO enables Qwen3-4B to start reasoning with high quality. As a result, the prefix tokens generated by Qwen3-4B with PPPO training can guide the reasoning process towards a desirable trajectory.

\textbf{Continuation accumulated reward enhances training effectiveness and stability.} By using Qwen3-4B as the backbone model, we implement PPPO with various generation times $G$. For each $G$, we train 4 times and report the average accuracies (\textit{avg@4}) and accuracy variances (\textit{var@4}) in Table~\ref{performance_2}. We can find that the single-sample reward strategy yields low accuracy and high variance. In contrast, the reward mechanism adopted by PPPO increases the average accuracy by up to 8.37\% and reduces the accuracy variance by up to 2.67\%. The reason is that our reward mechanism mitigates the randomness of single sampling through multiple explorations, which enables accurate evaluation for prefix tokens.

\begin{table}[t]
\centering
\begin{tabular}{c|c|ccc}
\toprule
Method            & Sample Strategy         & G  & avg@4 $\uparrow$  & var@4 $\downarrow$ \\  \midrule
\multirow{4}{*}{PPPO} & Single & 1  & 60.46  & 3.30         \\ \cmidrule{2-5}
                          & \multirow{3}{*}{Multiple}                       & 4  & 66.11  & 1.47         \\
                          &                        & 8  & 69.36  & 0.63         \\
                          &                        & 16 & 69.53  & 0.56         \\
\bottomrule
\end{tabular}
\caption{\label{performance_2}
   Results under different sampling times $G$.
  }
\end{table}

\begin{figure}[t]
\centering
\includegraphics[width=.95\linewidth]{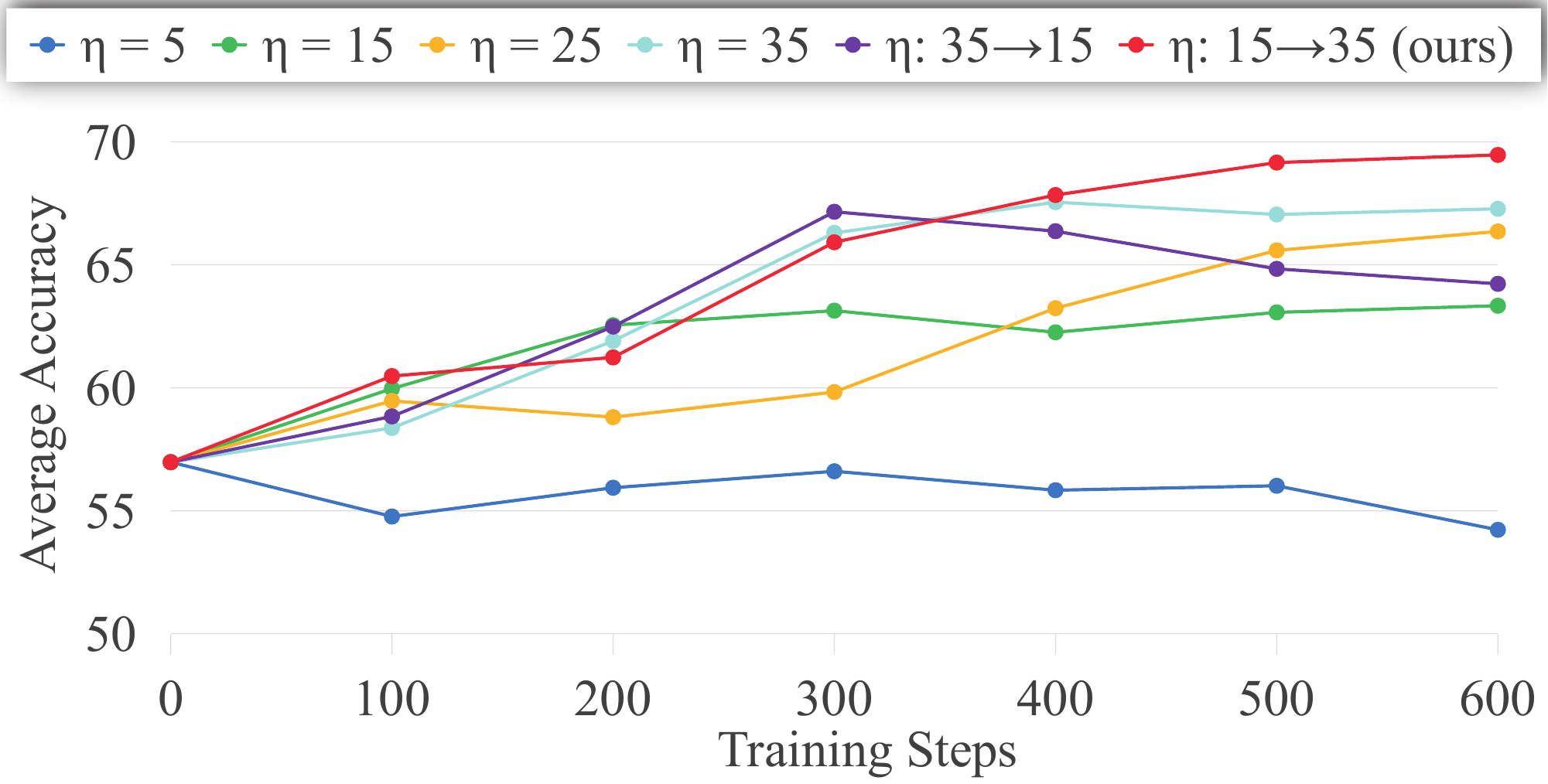} 
\caption{Training performance of PPPO with different prefix retention strategies. The backbone model is Qwen3-4B.}
\label{performance_3}
\end{figure}

\textbf{Progressive prefix retention strategy improves learning quality.} To demonstrate the superiority of our progressive prefix retention strategy, we compare it with 5 different prefix retention strategies. As shown in Figure~\ref{performance_3}, our progressive prefix retention strategy stabilizes the learning process of the model and increases the average accuracy by up to 12.49\%. Furthermore, compared with the strategy of ``$\eta=35$'', our strategy achieves better learning effectiveness (increasing the average accuracy by 2.19\% and reducing the number of training tokens by 8.83\%).

\section{Conclusion}

In this work, we introduce a critical yet underexplored phenomenon, Beginning Lock-in Effect, in LLM reasoning. It reveals that the quality of initial reasoning steps exerts a significant influence on the reasoning trajectory and outcome. By leveraging this insight, we propose a novel RLVR method termed Progressive Prefix-token Policy Optimization (PPPO). PPPO shifts the focus of RLVR training from uniform all-token optimization to a trajectory-specific optimization, where prefix tokens are optimized independently due to their disproportionate impact on reasoning processes. Comprehensive experiments across diverse reasoning tasks confirm that PPPO shows superior performance to various representative RLVR methods. Beyond methodological advancements, our findings highlight the importance of trajectory-specific optimization and indicate promising avenues for integrating cognitive insights into LLM research, which ultimately contributes to the development of more robust and human-aligned LLM systems.

\section*{Acknowledgements}

This research is supported by NSF of China (Nos: 62336003, 12371510), and Shanghai Municipal Science and Technology Major Project (No: 2025SHZDZX025G12)

\bibliography{aaai2026}

\end{document}